\documentclass{article}
\usepackage{amssymb}
\usepackage{graphicx}
\usepackage{floatrow}
\usepackage{caption}
\usepackage[table]{xcolor}
\usepackage{amsmath}
\usepackage{bm}
\usepackage{wrapfig}
\usepackage[normalem]{ulem}
\newfloatcommand{capbtabbox}{table}[][\FBwidth]
\usepackage{blindtext}
\definecolor{lightblue}{RGB}{242,250,253} 
\usepackage[final]{corl_2025} 
\usepackage[utf8]{inputenc}
\usepackage[T1]{fontenc}

\usepackage{tikz}
\usepackage{eso-pic}

\newcommand\CoRLaccepted{%
    \begin{tikzpicture}[remember picture,overlay]
        \node[anchor=north, yshift=-3.0em] at (current page.north) {%
            \textcolor[gray]{0.5}{\large\textrm{%
                \parbox{\textwidth}{\centering
                    This paper has been accepted for publication at the Conference on Robot Learning (CoRL), Seoul 2025. Please cite the final version: https://proceedings.mlr.press/v305/zeng25a.html%
                }%
            }}%
        };
    \end{tikzpicture}%
}

\AddToShipoutPictureBG*{\AtTextUpperLeft{\CoRLaccepted}}

\newcommand{\norm}[1]{\left\lVert#1\right\rVert}

\title{Decentralized Aerial Manipulation of a Cable-Suspended Load using \\ Multi-Agent Reinforcement Learning}

\author{
  \href{https://jackzeng-robotics.github.io/}{Jack Zeng}\textsuperscript{1,*},
  \href{https://andreumatoses.github.io/}{Andreu Matoses Gimenez}\textsuperscript{1},
  \href{https://www.eugenevinitsky.com/}{Eugene Vinitsky}\textsuperscript{2},
  \href{https://autonomousrobots.nl/people/}{Javier Alonso-Mora}\textsuperscript{1},
  \href{https://sihaosun.github.io/}{Sihao Sun}\textsuperscript{1,*} \\
  \textsuperscript{1}Delft University of Technology
  \textsuperscript{2}NYU Tandon School of Engineering \\
  *Corresponding authors: \texttt{jack-zeng@hotmail.com}, \texttt{s.sun-2@tudelft.nl}
}

\begin{document}
\maketitle

\begin{abstract}
    This paper presents the first decentralized method to enable real-world 6-DoF manipulation of a cable-suspended load using a team of Micro-Aerial Vehicles (MAVs). Our method leverages multi-agent reinforcement learning (MARL) to train an outer-loop control policy for each MAV. Unlike state-of-the-art controllers that utilize a centralized scheme, our policy does not require global states, inter-MAV communications, nor neighboring MAV information. Instead, agents communicate implicitly through load pose observations alone, which enables high scalability and flexibility. It also significantly reduces computing costs during inference time, enabling onboard deployment of the policy. In addition, we introduce a new action space design for the MAVs using linear acceleration and body rates. This choice, combined with a robust low-level controller, enables reliable sim-to-real transfer despite significant uncertainties caused by cable tension during dynamic 3D motion. We validate our method in various real-world experiments, including full-pose control under load model uncertainties, showing setpoint tracking performance comparable to the state-of-the-art centralized method. We also demonstrate cooperation amongst agents with heterogeneous control policies, and robustness to the complete in-flight loss of one MAV. Videos of experiments: \url{https://autonomousrobots.nl/paper_websites/aerial-manipulation-marl}

\end{abstract}

\keywords{Aerial Manipulation, Multi-Agent Reinforcement Learning,  Micro Aerial Vehicles} 


\section{Introduction}
\label{sec:introduction}
	
\begin{figure}[ht]
    \centering
    \includegraphics[width=1.0\linewidth]{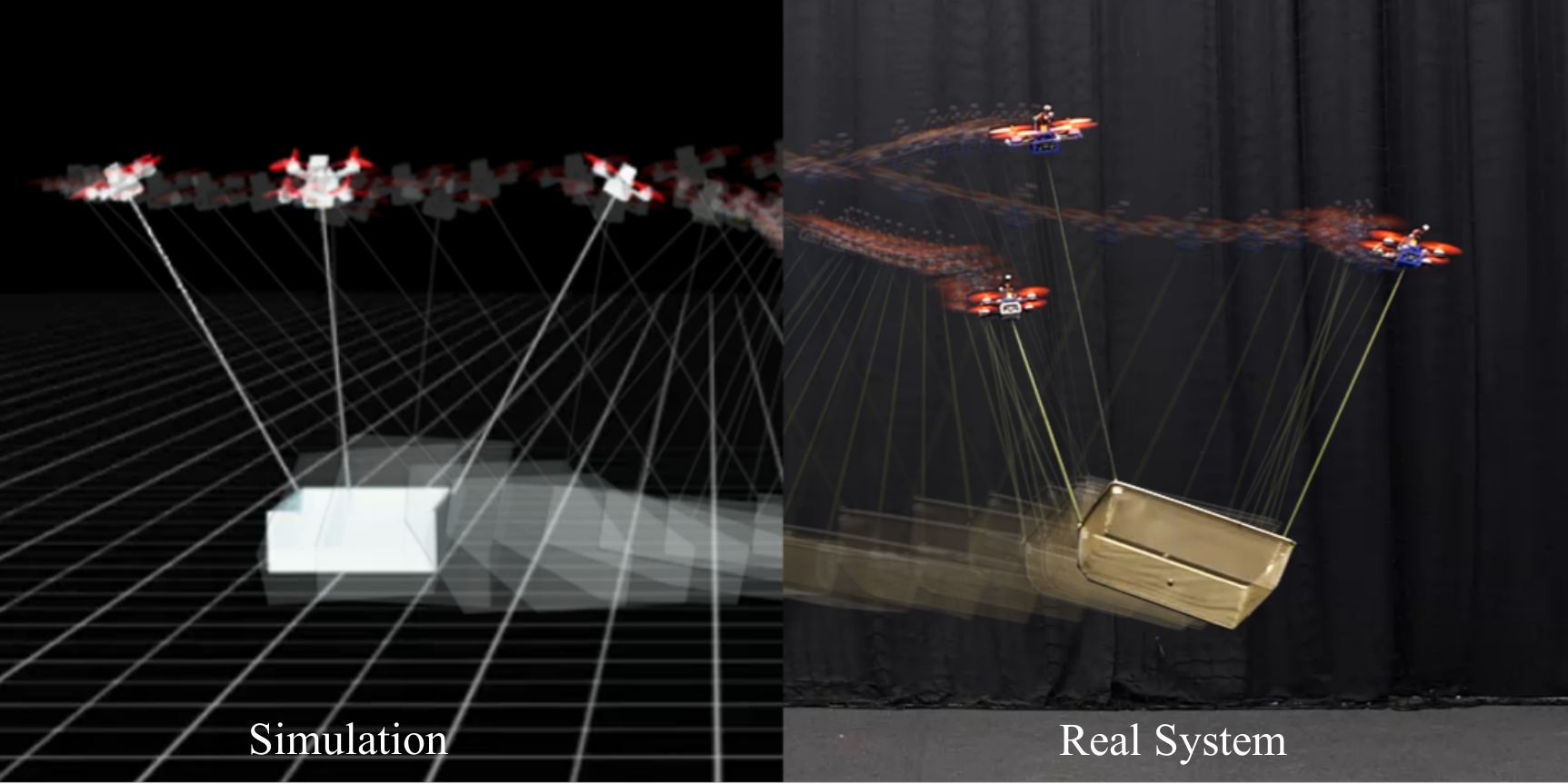}
    \caption{Multi-MAV lifting system performing full-pose control of a cable-suspended load. Left: simulation environment used to train the decentralized outer-loop control policy. Right: policy transferred to the real system.}
    \label{fig:multiliftsystem}
\end{figure}

Autonomous Micro Aerial Vehicles (MAVs) offer great capability for transporting slung loads to dangerous and remote locations~\cite{ barmpounakis_unmanned_2016}. 
While a single low-cost MAV has limited payload capacity, collaborative teams of MAVs can transport significantly heavier loads.
In addition, by connecting each MAV with the load at different points using tethers, the full pose of the load can be controlled by changing the position of the MAVs, yielding a cooperative cable-suspended manipulation solution, which shows great potential for aerial-based construction, inspection, and resecuring~\cite{sreenath_dynamics_2013, lee_geometric_2018, geng_load-distribution-based_2022, li_nonlinear_2023, sun2025agilecooperativeaerialmanipulation}.

To coordinate and control MAV fleets, the state-of-the-art method~\cite{sun2025agilecooperativeaerialmanipulation} employs a centralized framework that accurately captures the strong dynamical coupling between the MAVs and the suspended load. 
This ensures safety and stability while addressing the significant underactuation inherent to cable-suspended systems, preventing actuator saturations and reciprocal collisions. 
However, using centralized control strategies for such systems suffers from critical drawbacks: computational complexity tends to scale exponentially with the number of agents for many approaches, rendering real-time control infeasible for larger teams with a centralized scheme~\citep{sun2025agilecooperativeaerialmanipulation,bakule_decentralized_2012}. In addition, dependence on global state information and centralized communication is often impractical due to limits on sensors and communication bandwidth. A plausible solution, decentralization, remains an open challenge to effectively coordinate MAV fleets due to partial observability, limited communication bandwidth, and decision-making under strong dynamical coupling between agents while co-manipulating an object.

In this work, we present the first decentralized algorithm to achieve a real-world demonstrated full-pose manipulation of a cable-suspended payload using a team of MAVs.
Our method leverages multi-agent reinforcement learning (MARL) and \textbf{does not require any inter-agent communication}.
Instead, each agent only takes their own state and identity, the load pose, and the target load pose as observations.
We train the policy through MARL in a centralized training with decentralized execution (CTDE) paradigm using multi-agent proximal policy optimization (MAPPO)~\citep{yu_surprising_2022}. 
Each MAV learns to communicate implicitly through the load pose information.
To fill the sim-to-real gap in this highly dynamic cooperative task, we design the action space of the reinforcement learning (RL) policy as reference linear accelerations and body rates of the MAV and combine the RL policy with a low-level controller based on incremental nonlinear dynamic inversion (INDI)~\cite{smeur_adaptive_2016, tal_accurate_2021, sun_comparative_2022}.
The low-level controller follows the linear acceleration command with the body rate reference as the feedforward commands, ensuring agile and smooth control maneuvers during the cooperative manipulation.

Our method enables zero-shot transfer of the policy from simulation to real-world deployment to achieve full-pose control accuracy comparable to the state-of-the-art centralized controller~\cite{sun2025agilecooperativeaerialmanipulation}, and is deployed fully onboard.
In addition, experiments with real MAVs demonstrate that our method remains robust under load model uncertainties, operates effectively in heterogeneous agent settings where one MAV uses a different controller, and remains functional even when one of the MAVs completely fails.

\textbf{Our core contributions are as follows:}

\begin{itemize}
    \item The first method to achieve fully decentralized and onboard-deployed cooperative aerial manipulation in experiments with real MAVs, without any inter-agent communication.
    \item A novel action space design for MAVs manipulating a cable-suspended load, together with a robust low-level controller, enabling successful zero-shot sim-to-real transfer.
    \item First demonstration of robust full-pose control of the cable-suspended load under heterogeneous conditions and even under complete in-flight failure of an MAV.
\end{itemize}

\section{Related works}
\label{sec:related_works}
\textbf{Cooperative aerial manipulation} of a cable-suspended load typically embraces a centralized paradigm to consider the cable-load-MAVs system as a whole and requires global state observations to ensure safety and performance.
Early research on multi-MAV cable-suspended load problems often relied on model simplifications, such as assuming a quasi-static regime to ignore dynamic coupling effects~\citep{fink_planning_2011, michael_cooperative_2011, manubens_motion_2013, sanalitro_full-pose_2020}, which cannot address force-related constraints and perform dynamic motions. Another class of methods leverages system flatness~\cite{sreenath_geometric_2013} and dynamic equations to account for dynamic coupling effects.
An example is the cascaded scheme, which employs an outer-loop geometric controller to generate the commanded wrench for the load, distributes it as desired cable tensions, and executes it through inner-loop controllers of MAVs~\cite{lee_geometric_2018, li_cooperative_2021, li_rotortm_2024, wahba_efficient_2024}.
The outer-loop controller can be replaced by various approaches, such as inverse dynamics control~\cite{masone_shared_2021}, linear quadratic regulator~\cite{geng_load-distribution-based_2022}, and nonlinear model predictive control (NMPC)~\cite{li_nonlinear_2023}.
Recent work~\cite{sun2025agilecooperativeaerialmanipulation} leverages whole-body dynamics and NMPC to generate reference trajectories followed by an adaptive low-level controller, showing high agility and accuracy.

However, these centralized methods require exponentially higher computational budgets and communication burdens with the number of agents involved.
Therefore, decentralized controllers, such as distributed MPC~\citep{wehbeh_distributed_2020,wang_auto-multilift_2024} have been proposed and tested in simulation to address the problem with the computational issues. But these methods still require reliable inter-agent data transfer to obtain real-time states from other agents, which does not fundamentally solve the problems with limited communication bandwidth. 

\textbf{Multi-agent reinforcement learning} has been extensively studied for complex multi-agent systems, including cooperative scenarios~\citep{busoniu_comprehensive_2008, zhang_multi-agent_2021, sukthankar_cooperative_2017}. Beyond achieving expert-level performance in video games~\citep{vinyals_grandmaster_2019, openai_dota_2019}, MARL has been successfully applied to robotics, enabling decentralized control of multiple agents. For instance, researchers have leveraged MARL to develop cooperative strategies in robot football~\citep{liu_motor_2022, li_marladona_2024}, as well as multi-robot object manipulation with quadrupedal robots, including pushing~\citep{feng_learning_2024} and cable-based towing~\citep{chen2025decentralized}. Unlike our approach, these manipulation methods~\citep{feng_learning_2024, chen2025decentralized} rely on neighboring agent information through communication or onboard perception. In many cases, MARL is employed to optimize high-level task objectives while relying on mid- and low-level controllers for motor and sub-task execution, capitalizing on RL's ability to optimize a long-horizon task-level objective~\citep{song_reaching_2023}. 

Recent work by~\cite{gao_coohoi_2024} demonstrates MARL’s potential for cooperative object manipulation using simulated humanoids, relying solely on object bounding box information without explicit inter-agent communication. However, their approach depends on handcrafted reward functions that guide the humanoids toward predefined grasping points and walking behaviors. In MAV applications, MARL has been explored for tasks like swarming~\citep{batra_decentralized_2022}, but challenges remain due to the platform's agility, instability, and reliance on high-frequency, low-latency control~\citep{xingbootstrapping}.
Recently, MARL has shown potential for training multi-MAV lifting systems using global state observations~\cite{xu_omnidrones_2024}. However, a significant challenge remains to address the sim-to-real gap and partial observability, especially for the multi-MAV lifting system, where dynamic uncertainties are substantial due to complex aerodynamic disturbances and unknown cable tensions.

Our method effectively bridges this gap by leveraging multi-agent reinforcement learning (MARL) to achieve the first real-world demonstration of decentralized aerial manipulation, operating without global state observations or inter-agent communication. Furthermore, the method is deployed entirely onboard, enabled by its computational efficiency.

\section{Methods}
\label{sec:methods}
\begin{figure}[h]
    \centering
    \includegraphics[width=1.0\linewidth]{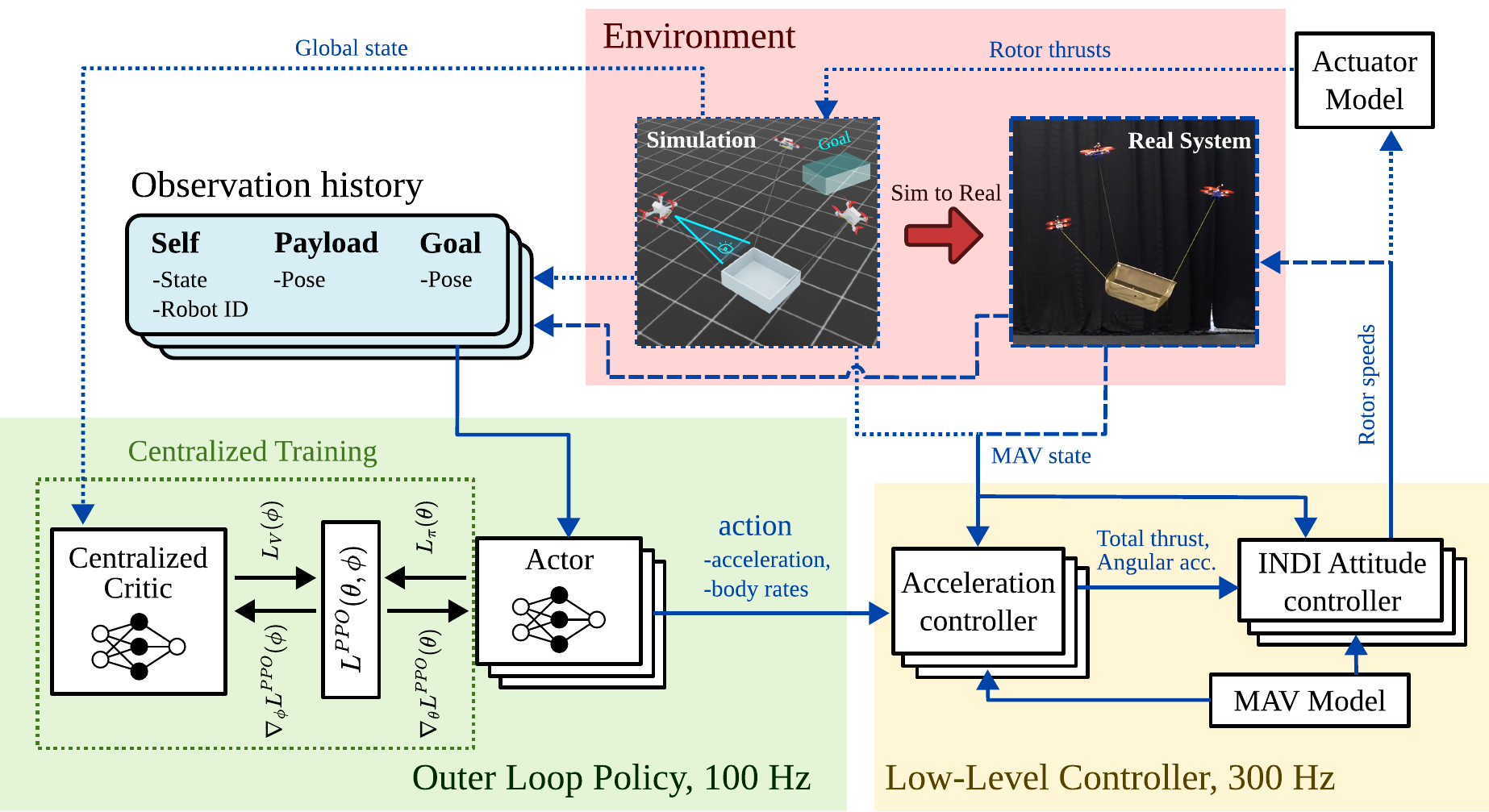}
    \caption{Overview of our method. Dotted lines indicate components only for training; dashed lines indicate those only for real-system deployment; solid lines for both. The training process involves the centralized critic (which observes the privileged global state), direct access to MAV states, and the actuator model that maps rotor speeds to thrust forces. Shared actors make decisions based on local observations, without access to other agents’ states. The output actions, namely acceleration and body rates, are tracked by a robust model-based low-level controller based on INDI. 
    }
    \label{fig:methodoverview}
\end{figure}

An overview of the full approach is shown in Figure \ref{fig:methodoverview}. Our method utilizes MARL to train an outer-loop control policy, which generates reference accelerations and body rates for the low-level controller in real-time based on local observations of the ego-MAV state, its robot ID, payload- and goal pose. The low-level controller, including an INDI attitude controller, tracks these references based on the MAV model and accelerometer measurements. The privileged full state is observed by the centralized critic during training, which is discarded at execution time. Collected experience is shared across actors to update the parameters of a shared policy. This enables training to be centralized while execution remains decentralized, allowing each agent to run the policy independently onboard after zero-shot transfer from simulation to the real world.

We model cooperative aerial manipulation as a decentralized partially observable Markov decision process (Dec-POMDP)~\citep{oliehoek2016concise} with a shared reward function. A Dec-POMDP is defined by $\mathcal{\langle I, S, A, O, P, R, \gamma \rangle}$, where $\mathcal{I}$ denotes the set of agents with the total number of agents being equal to $N$, $\mathcal{S}$ is the environment state, $\mathcal{A = }\{a_i\}_{i=1}^{N}$ is the joint action space of all agents, $\mathcal{O = }\{o_i\}_{i=1}^{N}$ is each agent's partial observation of the environment, $\mathcal{P:S\times A \rightarrow S}$ is the transition model,  $R :\mathcal{S\times A\times S \rightarrow \mathbb{R}}$ is the shared reward function and $\gamma$ is the discount factor. At each timestep $t$, the current state $\mathbf{s}_t \in \mathcal{S}$ transitions to a new state $\mathbf{s}_{t+1}$ based on the joint action $\mathbf{a}_t \in \mathcal{A}$ and the transition function $\mathcal{P}$. Each agent $i$ then receives the shared reward as feedback from the environment.

Our approach employs the CTDE paradigm~\citep{lowe_multi-agent_2017}, utilizing privileged global state information during training for the asymmetric centralized critic while relying solely on local observations for policy execution. Each agent $i$ has a policy $\pi_i : \omega_i(o_i) \rightarrow a_i$ that maps its local observation, processed through its observation function $\omega_i$, to an action $a_i$. We implement parameter sharing across agents (i.e., $\pi_i = \pi_j \ \forall i,j$), thus reducing $\pi_i$ to a homogeneous policy $\pi$. The set of observation functions for all agents can be denoted as $\Omega = \{\omega_i\}_{i=1}^{N}$. The final decentralized partially observable problem is thus defined by the tuple $\mathcal{\langle I, S, A, O,} \Omega \mathcal{, P, R, \gamma \rangle}$

\textbf{Observations and rewards}
The state of each MAV is given by $\bm{\mathit{x}}_i = \Bigl[\,
\bm{\mathit{p}}_{M,i},\,
\bm{\mathit{R}}_{M,i},\,
\bm{\mathit{v}}_{M,i},\,
\bm{\mathit{\omega}}_{M,i}
\Bigr]$, where $\bm{\mathit{p}}_{M,i} \in \mathbb{R}^3$ denotes the MAV's position, $\bm{\mathit{R}}_{M,i} \in \mathbb{R}^9$ is the vector composed of elements of its rotation matrix, $\bm{\mathit{v}}_{M,i} \in \mathbb{R}^3$ and $\bm{\mathit{\omega}}_{M,i} \in \mathbb{R}^3$ denote its linear and angular velocities. We use the subscript $i$ to denote the $i$-th MAV. The state of the load is given by $\bm{\mathit{x}}_L = \Bigl[\,
\bm{\mathit{p}}_L,\,
\bm{\mathit{R}}_L,\,
\bm{\mathit{v}}_L,\,
\bm{\mathit{\omega}}_L
\Bigr]$ where $\bm{\mathit{p}}_L \in \mathbb{R}^3$ denotes the load's position, $\bm{\mathit{R}}_L \in \mathbb{R}^9$ is the vector composed of elements of its rotation matrix, $\bm{\mathit{v}}_L \in \mathbb{R}^3$ and $\bm{\mathit{\omega}}_L \in \mathbb{R}^3$ denote its linear and angular velocities. The state of the goal relative to the payload is denoted by $\bm{\mathit{x}}_G = \Bigl[\,
\bm{\mathit{d}}_G,\,
\bm{\mathit{R}}_G
\Bigr]$
where $\bm{\mathit{d}}_G \in \mathbb{R}^3$ and $\bm{\mathit{R}}_G \in \mathbb{R}^9$ represent the goal position relative to the current load position and the vector composed of elements of its relative rotation matrix from the current load orientation to the goal orientation respectively. All quantities are described in the inertial world frame $\mathcal{F}_I$. The global state that is observable to the centralized critic during training is then denoted as:
\begin{equation}
\bm{\mathit{s}} = \Bigl[\,
\bm{\mathit{x}}_L,\,
\bm{\mathit{x}}_G,\,
\bm{\mathit{x}}_{M,1},\,
\bm{\mathit{x}}_{M,2},\,
\cdots, 
\bm{\mathit{x}}_{M,N}
\Bigr]
\end{equation}

Where $N$ is the total number of MAVs. The local policies' observation space only includes the load pose, relative goal terms, their own respective MAV state, and a one-hot vector $\bm{\mathit{e}}_i$ indicating their identity to enable role differentiation among homogeneous agents, as the policy network parameters are shared across all MAVs. The observation space for the $i$-th MAV is described as:
\begin{equation}
\bm{\mathit{o}}_i = \Bigl[\,
\bm{\mathit{p}}_L,\,
\bm{\mathit{R}}_L,\,
\bm{\mathit{x}}_G,\,
\bm{\mathit{x}}_{M,i},\,
\bm{\mathit{e}}_i
\Bigr]
\end{equation}

As the problem is partially observable, we use a history of observations by stacking the current and last 2 observations of the policy~\citep{hausknecht2015deep}. For a more detailed discussion on the history length, we refer the readers to Appendix \ref{appendix:history_length}.

We train the policies using MAPPO~\cite{yu_surprising_2022}, a model-free MARL algorithm that extends PPO~\citep{schulman_proximal_2017} with CTDE. The reward at time $t$, denoted as $r_t$, is defined as:
\begin{equation}
    r_t = r_t^{\mathrm{pos}} + r_t^{\mathrm{ori}} + r_t^{\mathrm{down}} + r_t^{\mathrm{act}} + r_t^{\mathrm{br}} + r_t^{\mathrm{thrust}}
\end{equation}

Where $r_t^{\mathrm{pos}}$ and $r_t^{\mathrm{ori}}$ are rewards to track the goal position and orientation for the load, $r_t^{\mathrm{down}}$ encourages the MAVs to aim their (proxy) downwash away from the load for stability against aerodynamic disturbances, $r_t^{\mathrm{act}}$ and $r_t^{\mathrm{br}}$ penalize action changes from the last time step and large body-rate outputs respectively for smoother flight, $r_t^{\mathrm{thrust}}$ penalizes outputting large thrusts which encourages energy efficiency. For a detailed reward formulation, we refer the readers to Appendix \ref{appendix:reward_formulation}.

\textbf{Action space and low-level controller}
To balance reliable sim-to-real transfer with sufficient control authority, the choice of action space is critical. Prior work in single MAV control demonstrates that high-level outputs (e.g., position or velocity) enhance robustness to disturbances and sim-to-real gaps but limit performance, whereas low-level outputs (e.g., snap) improve tracking precision at the cost of larger transfer discrepancies~\citep{kaufmann_benchmark_2022, eschmann_learning_2024}. To address this trade-off, we propose a mid-level action space in desired accelerations and body rates (ACCBR). This approach preserves adequate control capability while also being robust against uncertain disturbances and model mismatches from the cable-suspended load. 

The low-level controller converts the acceleration reference $\boldsymbol{a}_{i,\mathrm{ref}}$ from the outer-loop policy to the thrust direction command through the following acceleration controller:
\begin{equation}
\bm{\mathit{z}}_{i,\mathrm{des}} 
= \frac{\boldsymbol{a}_{i,\mathrm{ref}} - \boldsymbol{g} - \boldsymbol{f}_{i, \mathrm{ext}} / m_i}{\|\boldsymbol{a}_{i,\mathrm{ref}} - \boldsymbol{g} - \boldsymbol{f}_{i, \mathrm{ext}} / m_i\|},~~~~\boldsymbol{f}_{i, \mathrm{ext}} = \mathit{m}_i \boldsymbol{a}_{i,\mathrm{filtered}} - \boldsymbol{f}_{i,\mathrm{filtered}}
\end{equation}
where external forces $\boldsymbol{f}_{\mathrm{ext}}$, primarily due to the cable tensions, are estimated using the MAV mass $\mathit{m}_i$, filtered accelerometer measurements $\boldsymbol{a}_{i,\mathrm{filtered}}$ and collective thrust $\boldsymbol{f}_{i,\mathrm{filtered}}$ computed from a classical quadratic thrust model and filtered rotor speed feedbacks~\citep{sun_comparative_2022}.
The desired attitude command and the policy output body-rate command are then sent to the INDI attitude controller to generate rotor speed commands. We refer readers to~\citep{smeur_adaptive_2016, tal_accurate_2021, sun_comparative_2022} for further details on INDI. 

\textbf{Training setup}
We train our method completely in simulation and achieve zero-shot transfer to real-world experiments. The simulation environment is built using NVIDIA's Isaac Lab~\citep{mittal_orbit_2023}, and the MARL algorithms are modified from~\citep{serrano2023skrl}. Training was conducted on a consumer-grade RTX 3090 GPU and completed in 17 hours. The network architecture is a 4-layer MLP of size $[1024, 512, 256, 128]$ for both the shared policies and the centralized critic. The inputs to the network are normalized stacked observation histories with history size $H=3$. For a complete overview of training details, network and agent parameters, we refer the readers to Appendix \ref{appendix:training_config}.

The MAVs with a cable-suspended load spawn uniformly between $-1$ and $1$ in $xy$, $0.5$ and $1.5$ in $z$, with a random heading. The goal position is sampled from the same range, but also allows pitch and roll of $\pm45^\circ$. Despite sampling of the goal is limited to the predefined sets, the policy is still able to generalize and reach goal poses outside of it during execution.


\section{Experiments and Results}
\label{sec:experiments_results}

\subsection{Real-world experiments}
\begin{wrapfigure}{r}{0.6\textwidth}
\centering
\includegraphics[width=1.0\linewidth,height=0.5\linewidth]{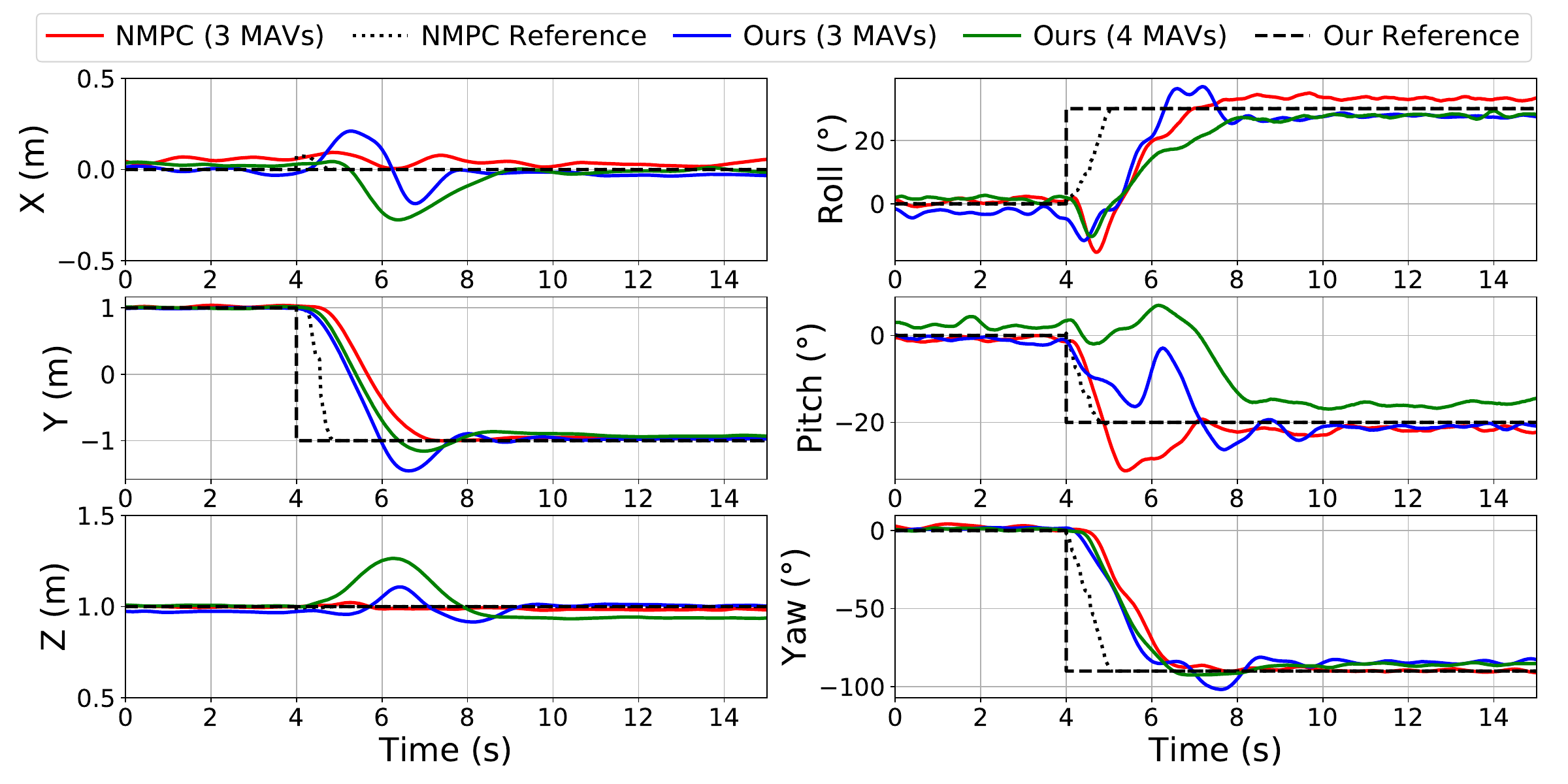}
\caption{Time series of pose tracking results comparing our method and a centralized NMPC method~\cite{sun2025agilecooperativeaerialmanipulation}. Our method also includes a setup with 4 MAVs.}
\label{fig:setpoint_tracking}
\end{wrapfigure}
\textbf{Setpoint tracking} Our real-world experiments demonstrate agile pose control of three MAVs with a cable-suspended load, tracking a 2~m displacement with (30\textdegree, -20\textdegree, -90\textdegree) attitude commands. We compare our decentralized method with the state-of-the-art centralized NMPC approach~\citep{sun2025agilecooperativeaerialmanipulation} in Figure \ref{fig:setpoint_tracking}. Despite being fully decentralized, our method achieves comparable tracking performance with positional and attitude RMSEs of 0.52~m (vs 0.45~m) and 22.93\textdegree~(vs 16.24\textdegree), respectively. Note that RMSE comparisons favor NMPC as it tracks a reference trajectory while we only track target poses, resulting in a larger RMSE in the transient area of the step command. The time-to-target (error $<\text{0.10}\,\text{m}/\text{10}^\circ$) is 6.84\,s for NMPC vs.\ 8.36\,s for ours, and the final displacement (RMSE) is 0.05\,m\,/4.02$^\circ$ for NMPC vs.\ 0.04\,m\,/5.78$^\circ$ for ours.
We also show successful pose control with 4 MAVs (without cable slack), resulting in tracking RMSEs of 0.92 m and 42.67\textdegree. The increased error, compared to the 3 MAV case, may be due to the system becoming overconstrained, which introduces more complex coordination and (cable) dynamics~\cite{li_rotortm_2024}. In terms of computational efficiency, we run the NMPC and our method onboard a Raspberry Pi~5 (2.4~GHz quad-core ARM Cortex-A76). Our method inferences in \textbf{6~ms} at 100~Hz, versus NMPC's \textbf{78~ms} at 10~Hz. Crucially, while NMPC's computation time grows exponentially with agent count, e.g., 174~ms and 267~ms for 5 and 6 agents respectively, our agent-independent approach maintains \textbf{constant computation time regardless of team size}.

\begin{figure}[]
    \centering
    \includegraphics[width=\linewidth]{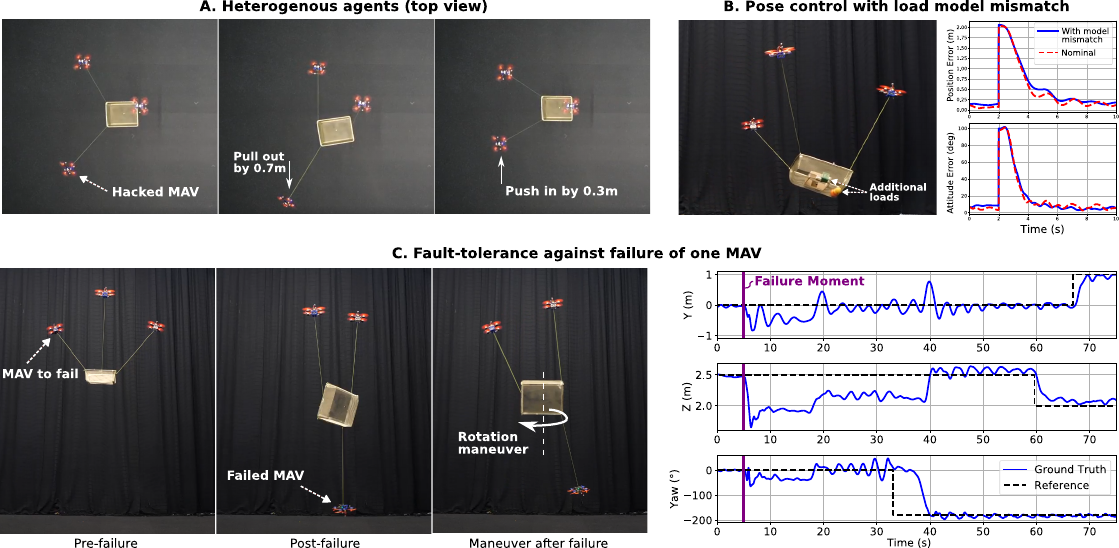}
    \caption{Real-world experiments. (A) Snapshot of the test with heterogeneous agents in which one MAV is manually controlled (hacked) to pull out and push in, and the other two MAVs counteract the interference of the hacked MAV. (B) Snapshot of the test where additional load is added to the original load, and the pose error with and without such model mismatch. (C) Snapshot of the case where one MAV fails in flight and the remaining two MAVs manage to control the load.}
    \label{fig:robustness_figures}
\end{figure}

\textbf{Robustness against load model mismatch} To evaluate robustness, we add objects (0.216~kg, 15.4\% of load mass) to the load, including four freely movable items that dynamically perturb both mass distribution and center of mass. Despite no inertia randomization during training, the system maintains strong tracking performance (0.63 m vs 0.60 m position RMSE; 26.93° vs 26.49° attitude RMSE. The low-level feedback controller automatically compensates for these disturbances, demonstrating inherent robustness to model uncertainties. Experimental results are shown in Figure~\ref{fig:robustness_figures}B.

\textbf{Heterogeneous agents} Although our policy is trained under the assumption of homogeneous agents, it remains effective when deployed with heterogeneous agents. In this experiment, we let the load hover at a fixed point.
Then we hacked one of the MAVs by replacing its RL policy with a model-based controller~\cite{sun_comparative_2022}, and provided it with different setpoints to observe the behavior of the other two MAVs controlled by the RL policy.
Specifically, we commanded the hacked MAV to move outwards on the y-axis by 0.7 m to pull the load away from the reference; we then commanded the hacked MAV to move inwards by 0.3 m to push it closer to the other two MAVs. Figure~\ref{fig:robustness_figures}A provides a snapshot of the experiments.
Since the policy is conditioned solely on the load pose and not on the states of the other agents, the two remaining MAVs utilizing the policy can compensate for load pose deviations from the reference. In contrast, the fully observable policy fails in these conditions due to the dependence on the states of all agents. Time series can be found in Appendix \ref{appendix:heterogeneous}.

\textbf{In-flight failure of one MAV} The effectiveness of our method with a heterogeneous agent setup and robustness against load model uncertainties also offers strong fault tolerance in the case of agent failure. In this experiment, we deliberately turned off the hacked MAV (one of the two on the same side). 
As a result, the load was controlled by the remaining two MAVs.
Note that with only two MAVs, the load orientation around the line joining the remaining two attachment points becomes unactuated.
Even worse, the failed MAV hangs underneath the load, leading to additional disturbances to the post-failure system.
Despite that, our method allows the other two MAVs to effectively control the remaining 5 DoFs of the load. We show that the system is still able to yaw by -180\textdegree~and is also able to maintain position control by flying 0.5 meters down along the z-axis and maneuvering along the y-axis by 1 meter. The tracking results and snapshots of the setup after the failure are seen in Figure \ref{fig:robustness_figures}C. As in the heterogeneous agent case, the remaining agents can compensate for the missing MAV since the policy operates independently of other agents' states, thereby avoiding unstable behavior in out-of-distribution scenarios. In contrast, the fully observable policy fails under these conditions due to its reliance on the states of all agents. Time series illustrating both scenarios can be found in Appendix \ref{appendix:failure}.

\subsection{Comparison among different action and observation spaces}

We compare our selected observation and action spaces with alternatives in simulation for safety. The Agilicious flight stack is used with the Gazebo simulator~\citep{koenig2004design} and RotorS~\citep{Furrer2016} plugins, which add sensor noise, aerodynamic disturbances, and system latencies in a ROS environment. All policies are trained for 1 billion environment steps (10 h) and evaluated 10 times in Gazebo.

\begin{figure}[ht]
    \centering
    \begin{floatrow}
        \ffigbox{%
            \includegraphics[width=0.9\linewidth]{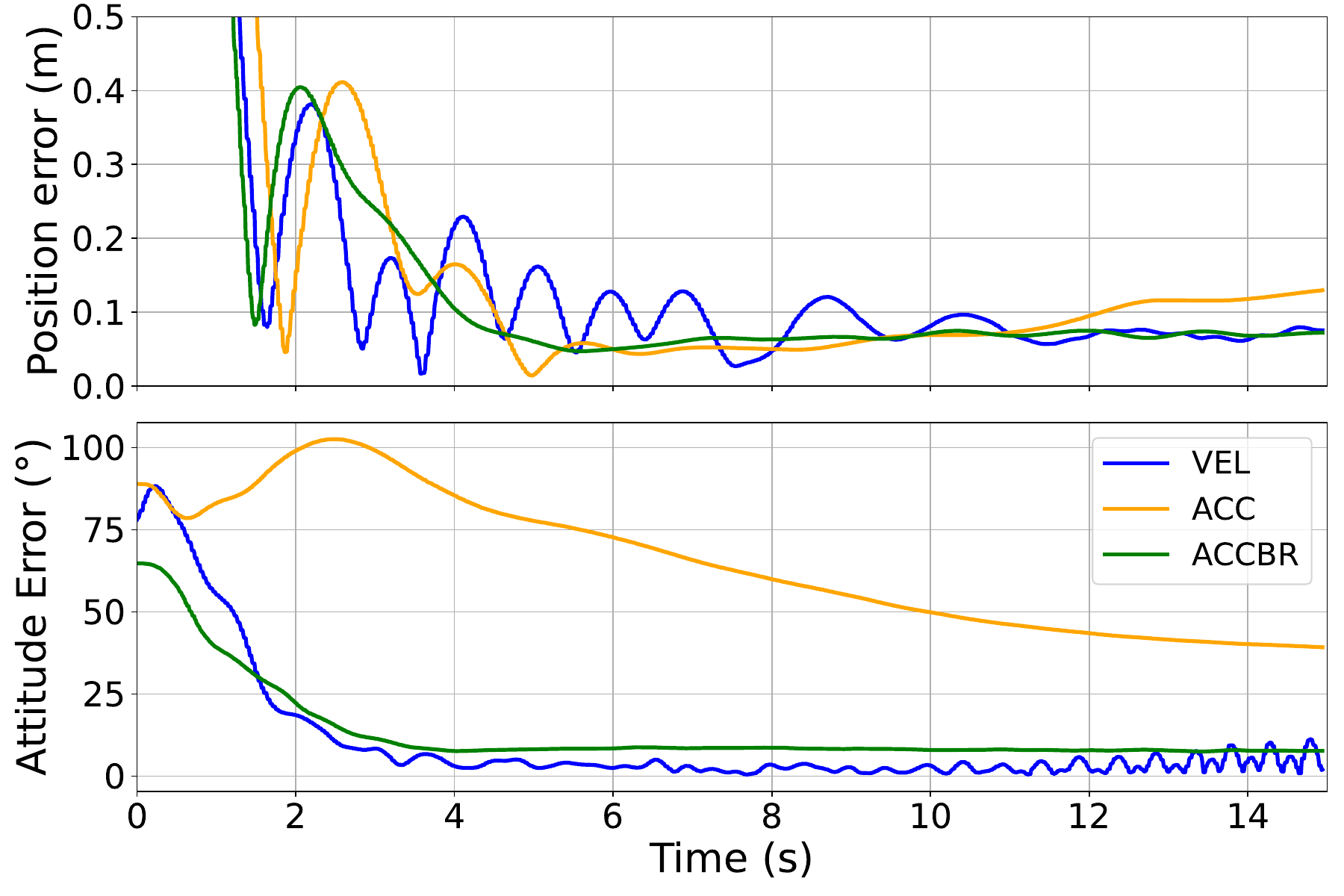}
        }{%
            \caption{Positional and attitude errors comparing different action spaces at test time in the Gazebo environment.}
            \label{fig:errors_action_ablation}
        }

        \capbtabbox{%
            \raisebox{2.2cm}{
                \small
                \setlength{\tabcolsep}{4pt}
                \begin{tabular}{c|cc}
                    \hline
                    \rowcolor{white}
                    \textbf{Action space} & \textbf{Pos RMSE} & \textbf{Att RMSE} \\
                    \hline
                    \rowcolor{lightblue}
                    ACCBR & 0.64 $\pm$ 0.00 & 33.87 $\pm$ 0.91 \\
                    \rowcolor{white}
                    CTBR$^*$ & NaN & NaN \\
                    \rowcolor{lightblue}
                    ACC & \textbf{0.54 $\pm$ 0.00} & 87.89 $\pm$ 1.85 \\
                    \rowcolor{white}
                    VEL & 0.56 $\pm$ 0.06 & \textbf{25.74 $\pm$ 1.49} \\
                    \hline
                    \multicolumn{3}{l}{\footnotesize $^*$Not able to take off} \\
                    \hline
                \end{tabular}
            }
        }{%
            \caption{Pose tracking RMSEs of different action spaces at test time in the Gazebo environment.}
            \label{tab:RMSE_action}
        }
    \end{floatrow}
\end{figure}

\textbf{Action space} 
We compare the ACCBR action space with three alternatives: velocity (VEL), linear acceleration (ACC), and collective thrust with body rates (CTBR). The ACCBR, VEL, and ACC outputs all utilize the same low-level controllers, which compensate for disturbances such as aerodynamic forces and cable tension. In contrast, CTBR outputs feed directly into the INDI attitude controller without additional disturbance compensation. 

The RMSE results in Table \ref{tab:RMSE_action} demonstrate that the VEL action space achieves the best performance, followed by ACCBR, while ACC fails to track the load orientation accurately. Notably, the widely used CTBR approach~\citep{kaufmann_champion-level_2023, song_reaching_2023} fails to learn effectively. Since CTBR directly commands collective thrust without leveraging the proposed low-level controller's disturbance compensation, we hypothesize that the unpredictable cable forces exerted on each MAV make the learning process prohibitively difficult, as there are no cable force sensors mounted for both training and evaluations. 


However, while VEL achieves lower RMSE, Figure~\ref{fig:errors_action_ablation} shows it causes \textbf{hazardous oscillations}. ACCBR offers more stable hovering despite higher initial errors, making it safer and preferable for stability-critical tasks like inspection or delivery.

\begin{wrapfigure}{r}{0.4\textwidth}
    \centering
    \includegraphics[width=\linewidth]{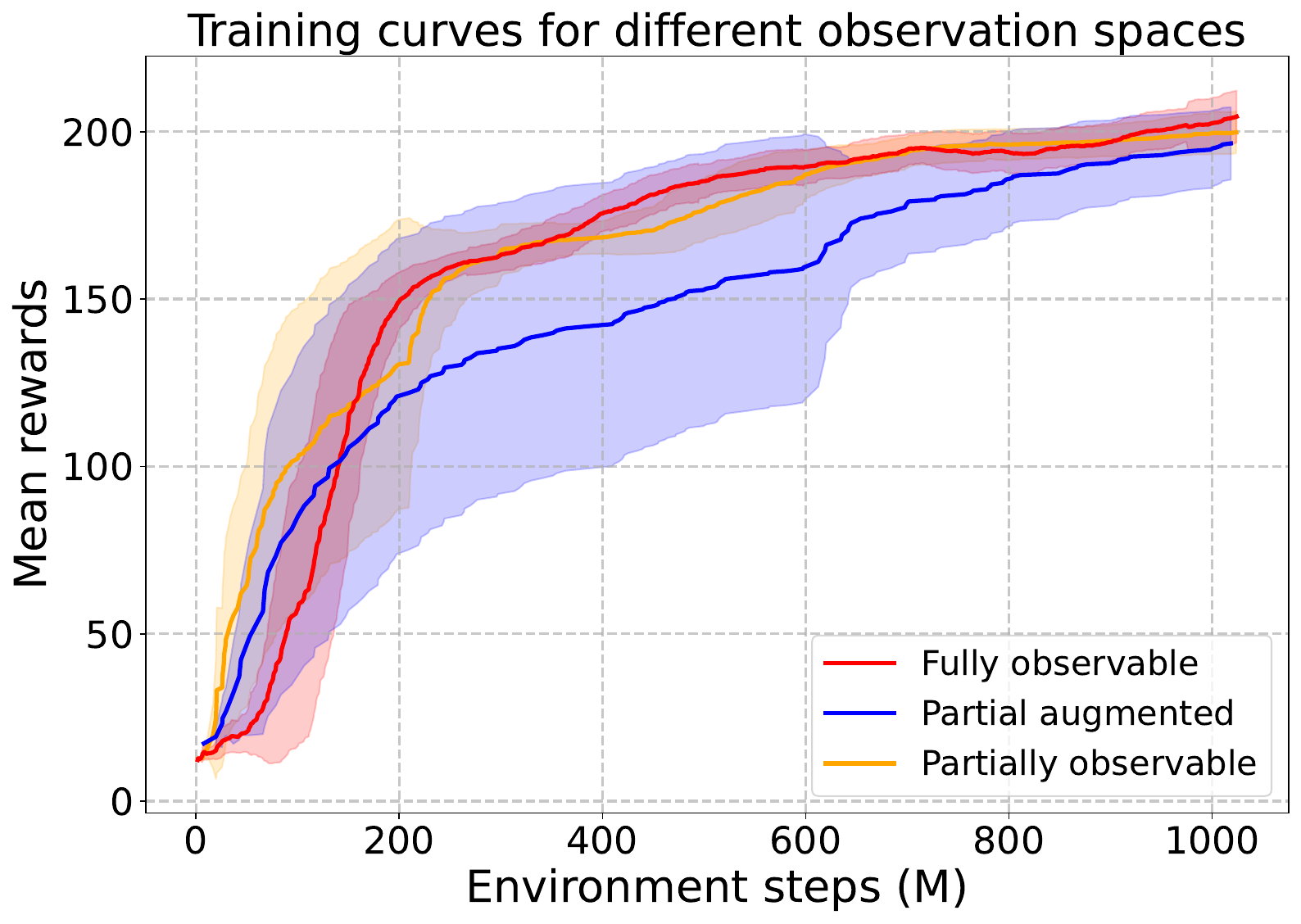}
    \caption{Training curves of fully observable, partial augmented, and partially observable observation spaces.}
    \label{fig:training_curves_obs}
\end{wrapfigure}

\textbf{Observation space} To benchmark the decentralized policy's performance, we compare three observation space cases: (1) the fully observable case with global state $\bm{s} = [\bm{x}_L, \bm{x}_G, \bm{x}_1, \bm{x}_2, \bm{x}_3]$, (2) an augmented partial observability case where each MAV $i$ also receives the load twist and other MAVs' positions ("Partial augmented") $\bm{o}_i = [\bm{x}_L, \bm{x}_G, \bm{p}_{j_1}, \bm{p}_{j_2}, \bm{x}_i, \bm{e}_i]$ with $\bm{p}_{j_1}, \bm{p}_{j_2}$ representing the neighboring agents' positions, and (3) the partially observable case. 
For partially observable cases, we include observation histories ($H=3$) to improve state estimation and decision-making under uncertainty~\citep{hausknecht2015deep}. Figure \ref{fig:training_curves_obs} reveals comparable convergence across all configurations, indicating that load pose alone serves as a sufficient statistic for implicit MAV coordination, while the full global state contains redundant elements. 


\section{Conclusion}
\label{sec:conclusion}
We introduced a decentralized method using MARL that allows for full-pose control of a cable-suspended load using three MAVs without any inter-MAV communication or neighboring MAV information.
The policy is computationally tractable and executes entirely onboard. We proposed a novel action space of accelerations and body rates (ACCBR) along with a robust low-level controller and showcase zero-shot transfer from simulation to real-world deployment. Extensive testing with real MAVs shows that the setpoint tracking performance of our method is comparable to that of the state-of-the-art centralized NMPC~\citep{sun2025agilecooperativeaerialmanipulation}, despite being fully decentralized and having significantly lower computation time. Our method demonstrates robustness against unknown disturbances, heterogeneous agents, and even the complete in-flight failure of one MAV. We attribute this resilience to two key factors: 1) closed-loop reference tracking by the low-level controller, which maintains stability despite perturbations, 2) decentralized policy independence, where local agents operate without dependence on neighboring states, preventing cascading failures. Our work shows promising results to enable scalable and robust cooperative aerial manipulation with minimal onboard sensing and no internal communications required.


\section{Limitations}
\label{sec:limitations}
Our method requires pose measurement of the load, which is not often practical beyond lab environments. In our experiment, we require an external motion capture system to provide high-frequency load pose measurement. For future real-world outdoor deployment, onboard sensing (e.g., a downward-facing camera for load pose estimation and SLAM for MAV localization) would be necessary. This would introduce new challenges, such as observation delays, imperfect state estimates, sensor noise, and different reference frames for the load and MAVs that require alignment and synchronization. Additionally, our current framework does not address obstacle avoidance, as we assume collision-free paths to the goal—an unrealistic assumption in unstructured environments. Future work will focus on integrating a robust perception stack and obstacle avoidance capabilities.

\clearpage

\acknowledgments{The authors would like to thank Dr. Yunlong Song, Dr. Dennis Benders, and Shlok Deshmukh for the insightful discussions, and Maurits Pfaff and Kseniia Khomenko for their help with the experiments.
This work is funded by the Dutch Research Council (NWO) under Grant 20256 for the project "Accurate Aerial Manipulation under Uncertainties" and by the European Union under the ERC grant INTERACT, 101041863. Views and opinions expressed are however those of the author(s) only and do not necessarily reflect those of the European Union or the European Research Council Executive Agency. Neither the European Union nor the granting authority can be held responsible for them.
}


\bibliography{reference} 

\clearpage
\appendix
\section{Appendix}
\subsection{Experimental setup}
\textbf{Real-world evaluation setup}
We evaluate our method in real-world experiments. Our experiment includes 3 MAVs built based on the Agilicious~\citep{foehn_agilicious_2022} flight stack. Each MAV is connected to a basket-shaped payload with 1-meter cables at three distinct locations. The MAVs weigh 0.6kg, and the payload weighs 1.4 kg.
We conduct the experiment in an indoor flight space with motion capture systems. We attach motion capture markers to the MAVs and the payload to measure their positions and orientations and distribute them to each MAV through ROS at 100 Hz. The trained policy and low-level controllers are deployed onboard each MAV. The policy is inferred at 100 Hz to send acceleration and body-rate commands. The low-level controller is executed at 300 Hz to generate rotor speed commands. 

\subsection{Heterogeneous agents time series} \label{appendix:heterogeneous}

\begin{figure}[ht]
    \centering
    \includegraphics[width=1.0\linewidth]{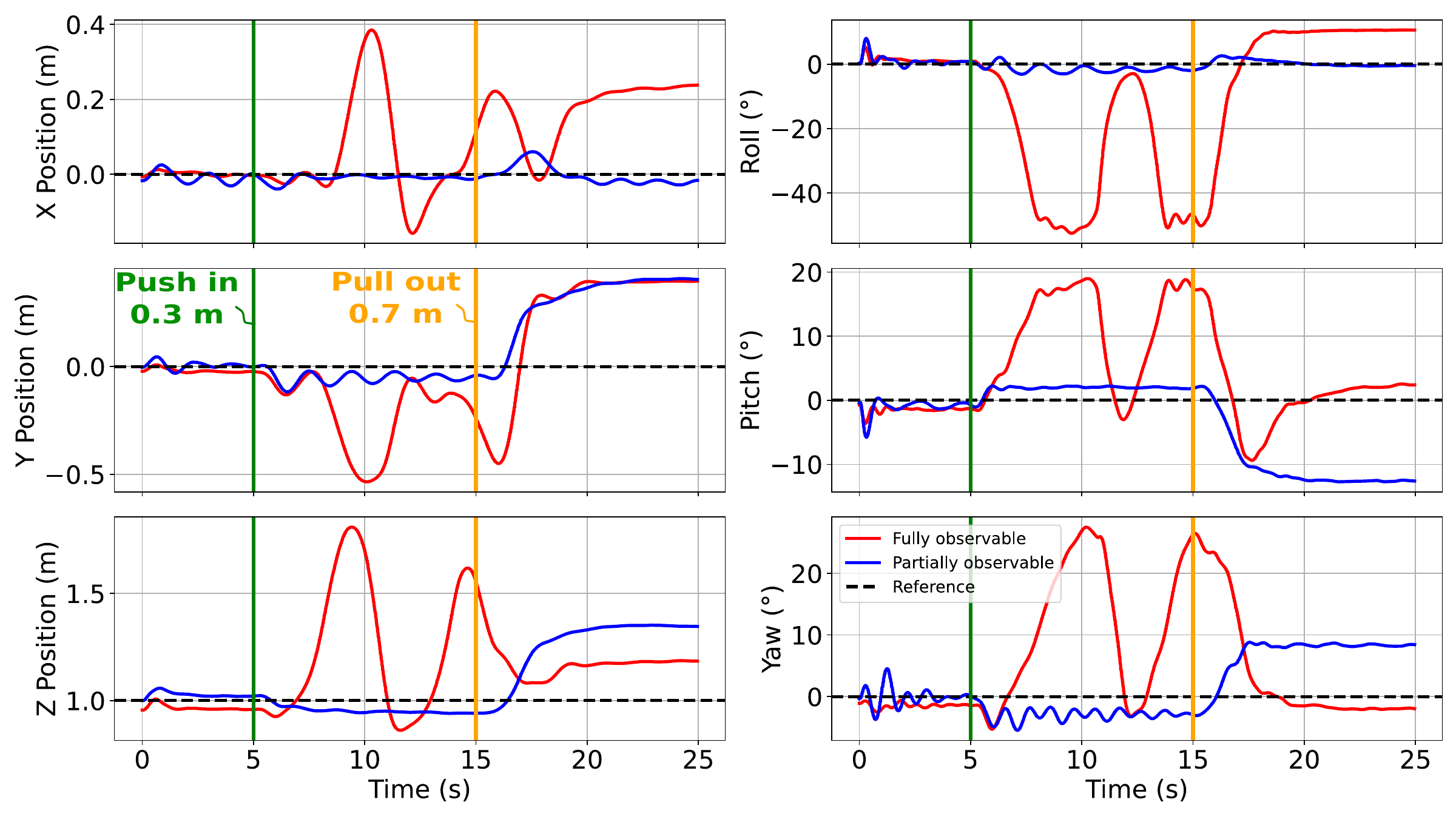}
    \caption{Time series of the load pose in the heterogeneous agents scenario, comparing the performance of the partially observable policy and the fully observable policy. The time points at which control commands are issued to push the load inward by 0.3 m relative to the desired policy position, or to pull it outward by 0.7 m, are indicated in green (push-in) and orange (pull-out), respectively.}
    \label{fig:heterogeneous_appendix}
\end{figure}

Figure \ref{fig:heterogeneous_appendix} compares the performance of partially observable and fully observable policies in the heterogeneous agents scenario. The partially observable policy, being independent of other agents' states, allows the unaffected MAVs to compensate for the hacked agent, maintaining system stability. In contrast, the fully observable policy—which relies on neighboring agents' states—performs worse, exhibiting larger tracking errors (0.42 m vs. 0.28 m in position, 30.08 degrees vs. 8.88 degrees in attitude) and large oscillations during the inward push.

\subsection{In-flight failure of one MAV time series} \label{appendix:failure}

\begin{figure}[ht]
    \centering
    \includegraphics[width=1.0\linewidth]{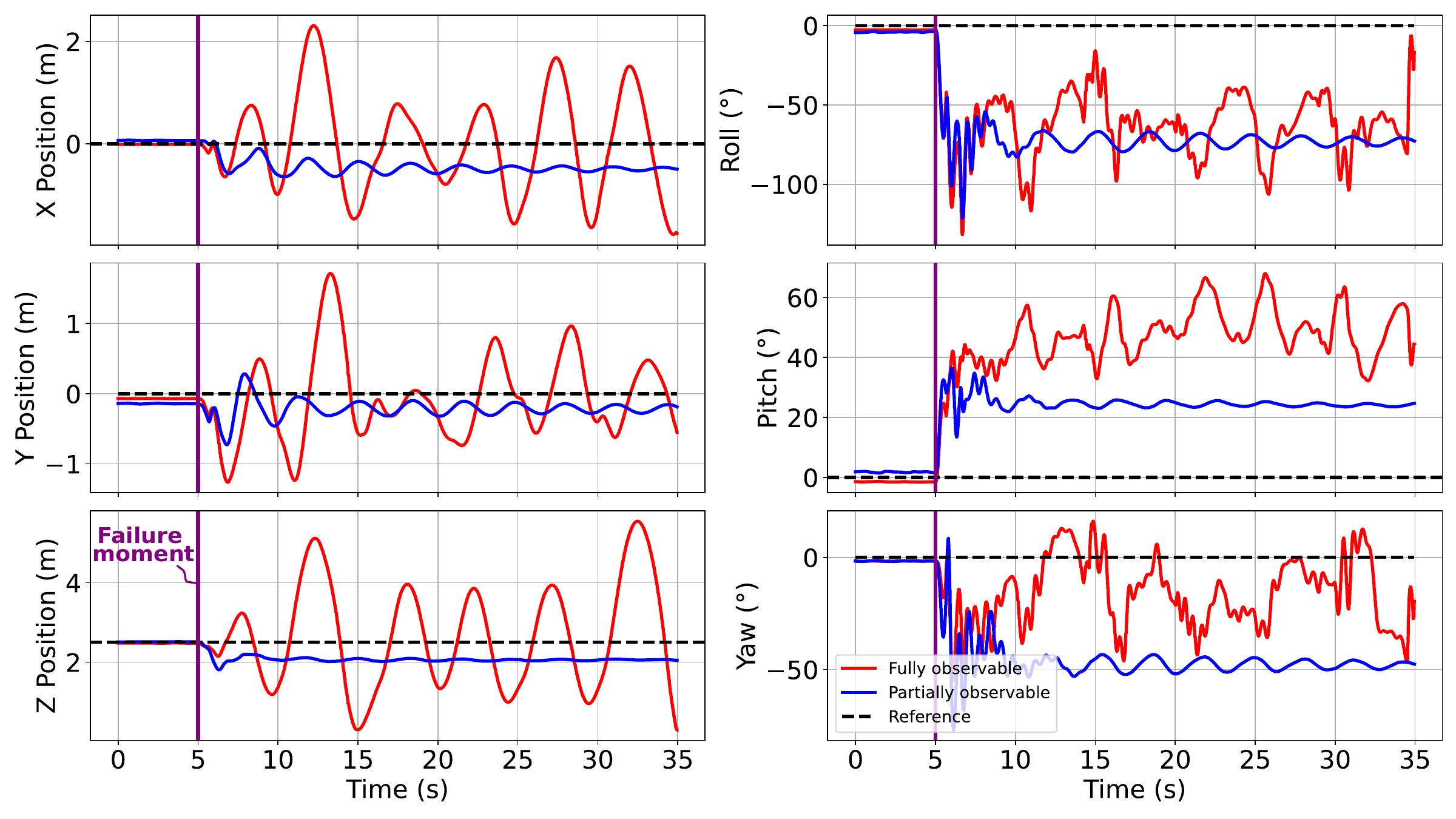}
    \caption{Time series of load pose in the in-flight failure of one MAV case without sending any commands, comparing a partially observable policy vs a fully observable policy. The thick purple line indicates the moment the MAV fails.}
    \label{fig:failure_appendix}
\end{figure}

\begin{figure}[h!]
    \centering
    \includegraphics[width=1.0\linewidth]{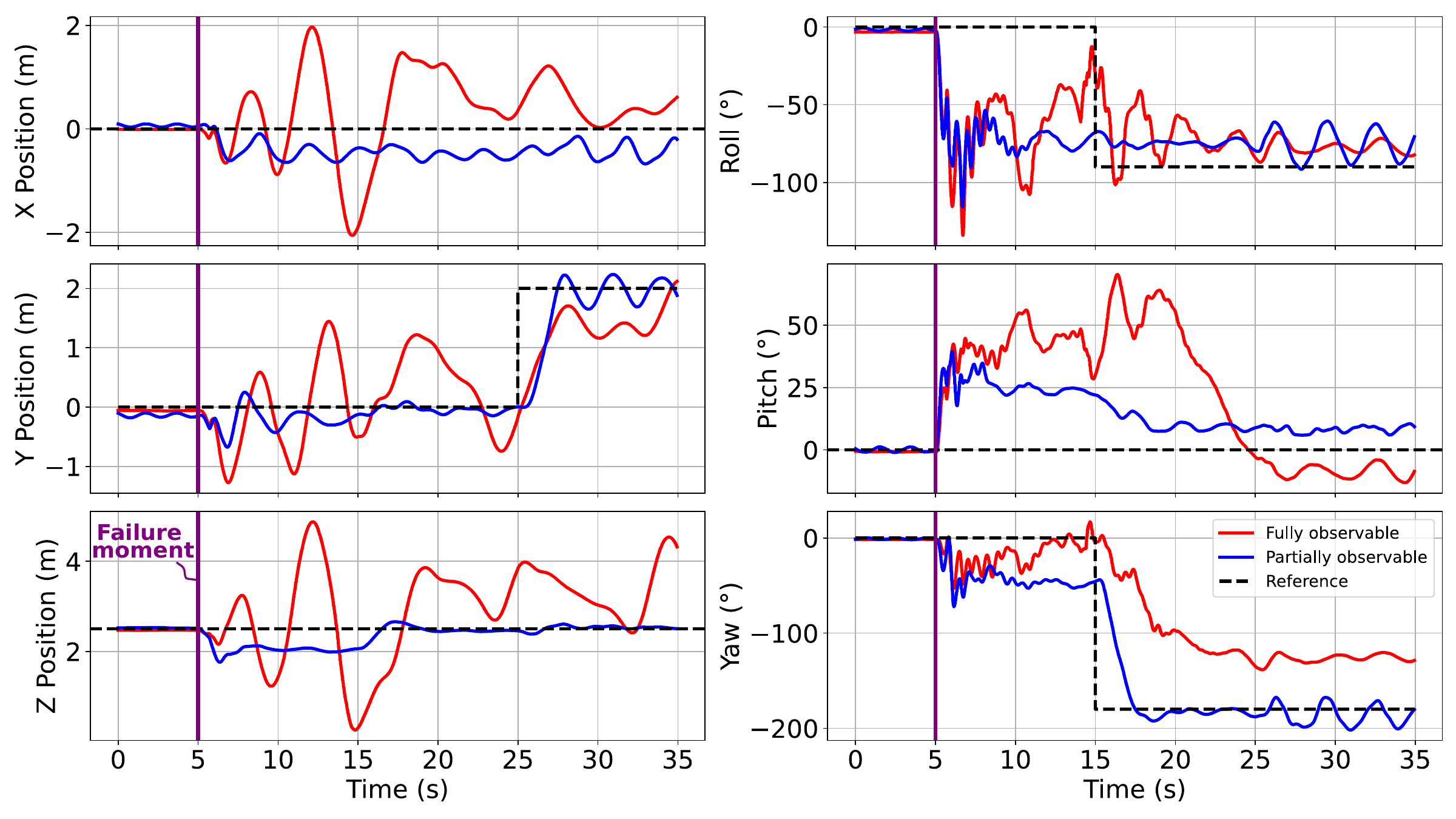}
    \caption{Time series of load pose in the in-flight failure of one MAV case, comparing a partially observable policy vs a fully observable policy. An attitude command is sent after 10 seconds and a positional command after 20 seconds. The thick purple line indicates the moment the MAV fails.}
    \label{fig:failure_command_appendix}
\end{figure}

Figures \ref{fig:failure_appendix} and \ref{fig:failure_command_appendix} show the tracking performance of the partially observable and fully observable policies following an in-flight failure of one MAV. Figure \ref{fig:failure_appendix} corresponds to the scenario in which no additional command inputs are issued, whereas Figure \ref{fig:failure_command_appendix} corresponds to the scenario in which new attitude and position commands are introduced at $t = 15$ s and $t = 25$ s. In both scenarios, the partially observable policy successfully compensates for the MAV failure. In contrast, the fully observable policy exhibits strong oscillatory behavior, causing the suspended MAV to repeatedly crash to the ground. When new pose commands are sent, the fully observable policy fails to track them accurately, whereas the partially observable policy is still able to track 5 DoF. This results in larger tracking errors for the fully observable policy, which incurs position and attitude root-mean-square errors of 1.50 m and 73.37 degrees, respectively, compared to 0.67 m and 50.31 degrees for the partially observable policy. The robustness of the partially observable policy is attributed to its independence from the states of neighboring agents, which helps prevent cascading failures.

\subsection{Trajectory tracking}
\begin{figure}[ht]
    \centering
    \includegraphics[width=0.95\linewidth]{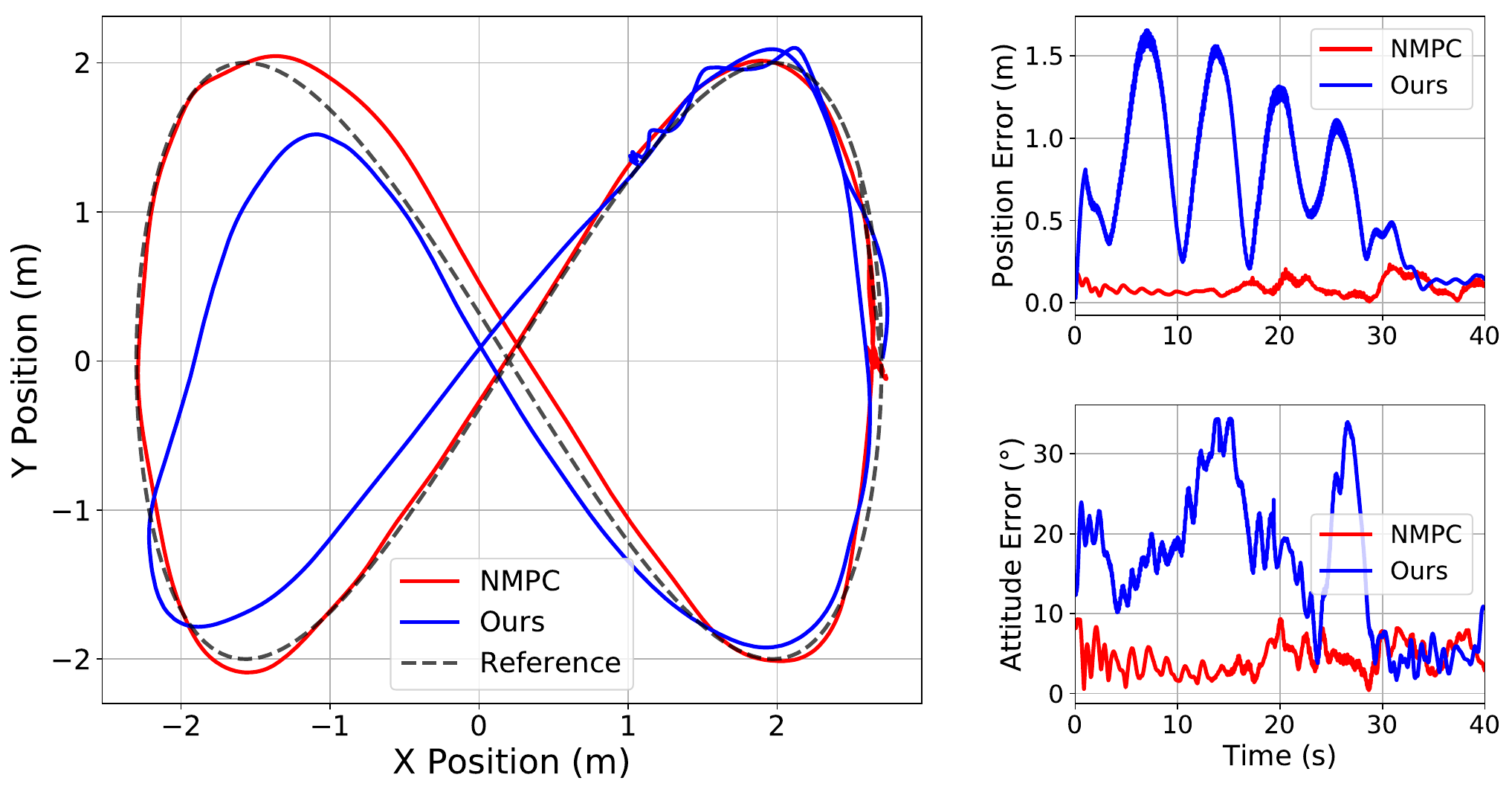}
    \caption{Comparison of our method, which is not trained for trajectory tracking, against the centralized NMPC in \cite{sun2025agilecooperativeaerialmanipulation}. \textbf{Left}: top view of the flight path of the center of mass of the load while tracking a figure-eight trajectory with a maximum velocity of 1 m/s and maximum acceleration of 0.5 m/s$^2$. \textbf{Right}: position (top) and attitude (bottom) tracking errors time series.}
    \label{fig:trajectory_tracking}
\end{figure}

Although our method is not trained for trajectory tracking, we evaluate its trajectory tracking capabilities against that of the centralized NMPC~\cite{sun2025agilecooperativeaerialmanipulation} in Figure \ref{fig:trajectory_tracking}. The reference trajectory is a figure-eight trajectory with a maximum velocity of 1 m/s and a maximum acceleration of 0.5 m/s$^2$. It is worth noting that our method only considers the reference pose information, while the NMPC also takes velocity information from the reference trajectory into account. For future specialized trajectory tasks, incorporating higher-order information such as velocity, as well as future reference points~\cite{mayer2021proximal} into the observations would significantly improve tracking performance and make for a fairer comparison. Nonetheless, our method is able to successfully track the figure-eight trajectory, albeit with a high tracking error. Our method achieves positional and attitude RMSEs of 0.82 m (vs 0.10 m), and 18.22 degrees (vs 4.80 degrees).

\subsection{Performance without centralized critic}

\begin{wrapfigure}{r}{0.48\textwidth}
    \centering
    \includegraphics[width=\linewidth]{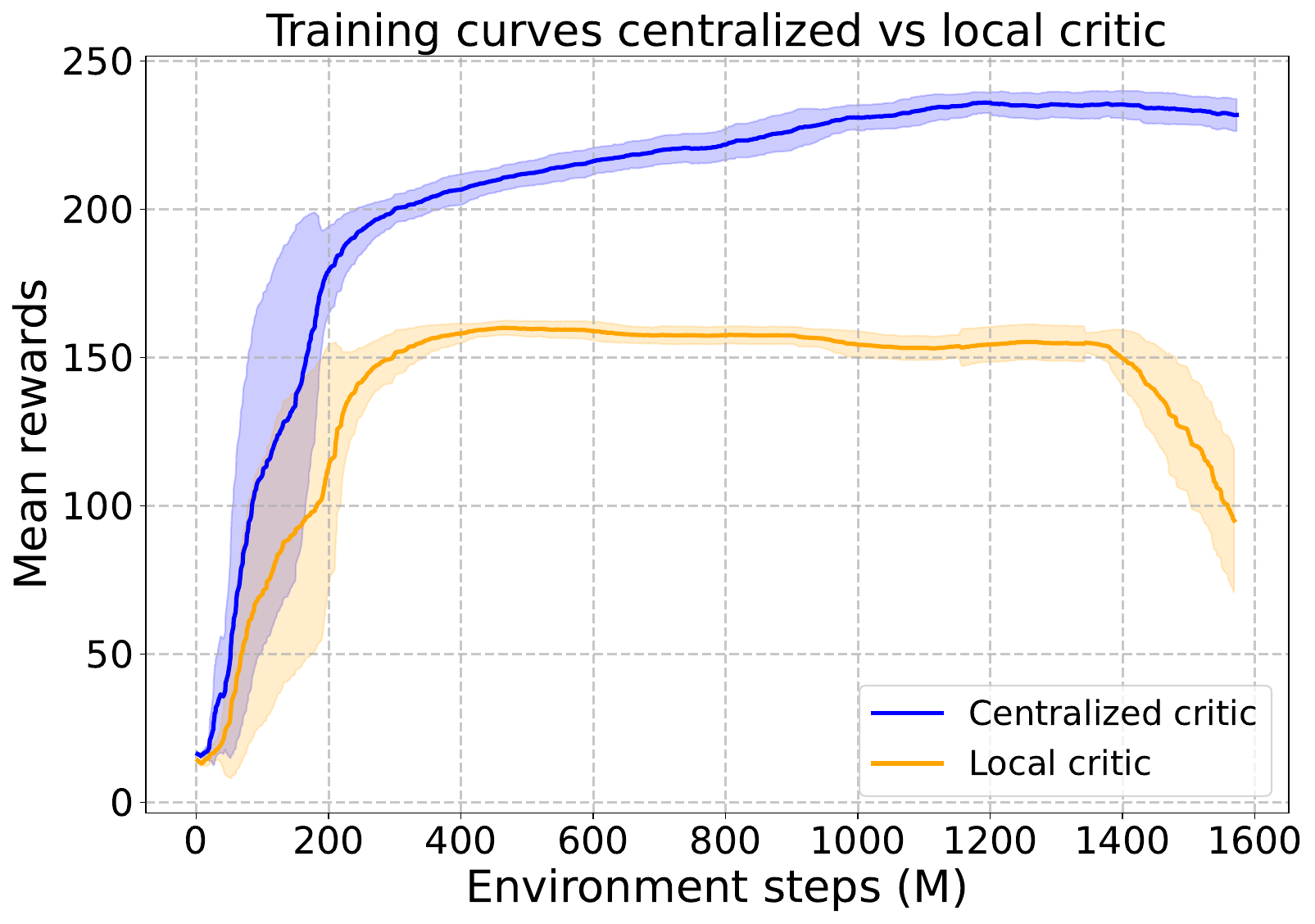}
    \caption{Training curves using a centralized critic vs using a local critic.}
    \label{fig:training_curves_critic}
\end{wrapfigure}

To assess the impact of using a centralized critic with access to privileged global state information, we compare its performance against a policy trained with a shared local critic. The local critic has access only to local observations, which are the same as those available to the actor. The training curves in Figure \ref{fig:training_curves_critic} show that the setup with the local critic fails to converge to the same performance as with the centralized critic, and even collapses at the end. Specifically, the policy with the local critic fails to learn the position and orientation rewards effectively. We hypothesize that access to global state information allows the centralized critic to produce more accurate value estimates, which can indirectly support more effective credit assignment during learning~\cite{foerster2018counterfactual}, thereby improving task performance.

\subsection{Performance with different history lengths} \label{appendix:history_length}

We compare the performance of the partially observable policy with different history lengths in the observation space. $H=1$ means that the history only contains the observations of the current timestep (no previous observations). All policies are trained on a limited budget of 2 billion environment steps and are evaluated in the Gazebo environment.

\begin{figure}[ht]
\floatsetup[figure]{valign=c}
\begin{floatrow}
    \ffigbox[0.50\textwidth]{
        \includegraphics[width=\linewidth]{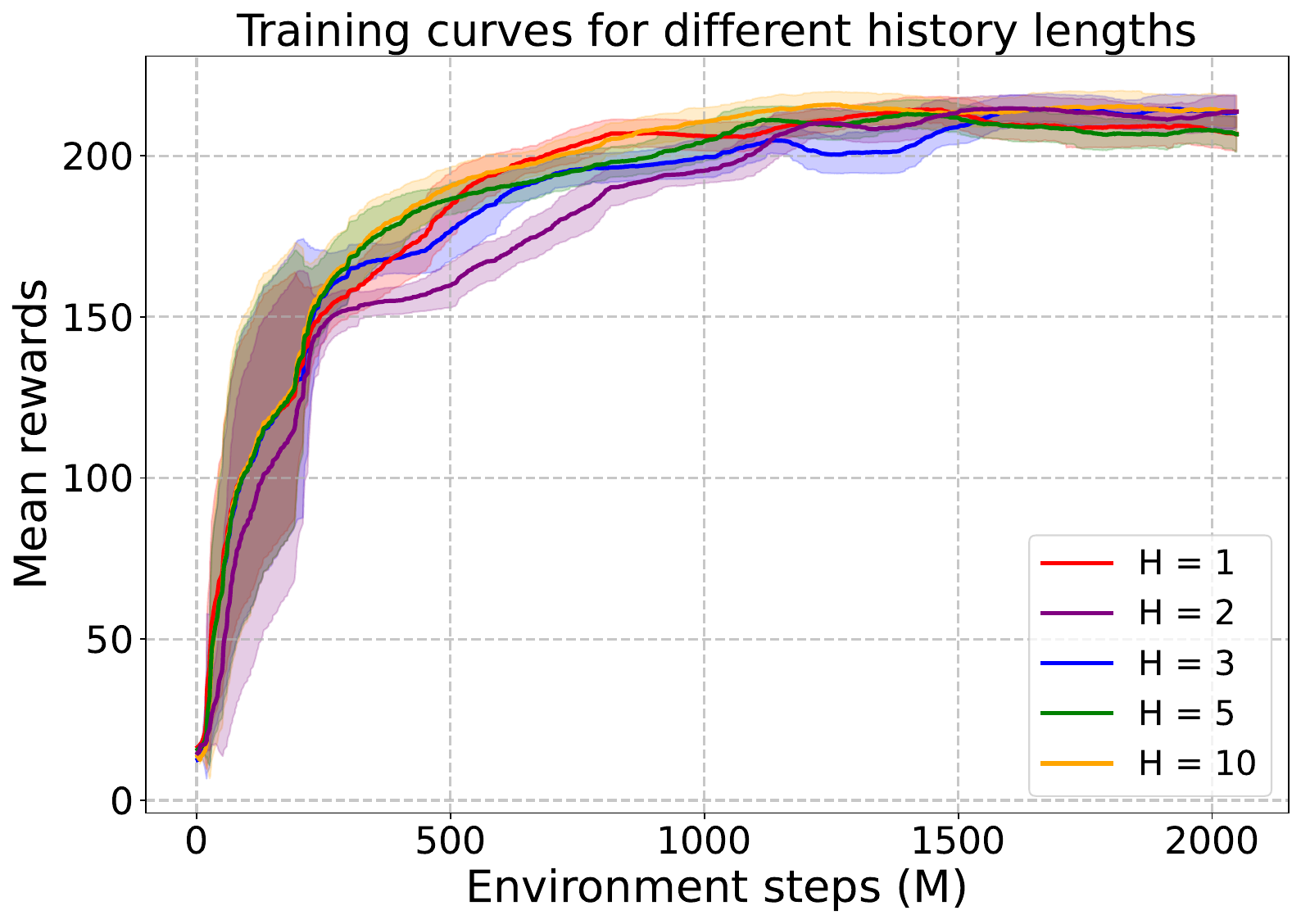}
    }{
        \caption{Training curves comparing different history lengths for the partially observable policy.}\label{fig:training_curves_history}
    }
    \ffigbox[0.50\textwidth]{
        \includegraphics[width=\linewidth]{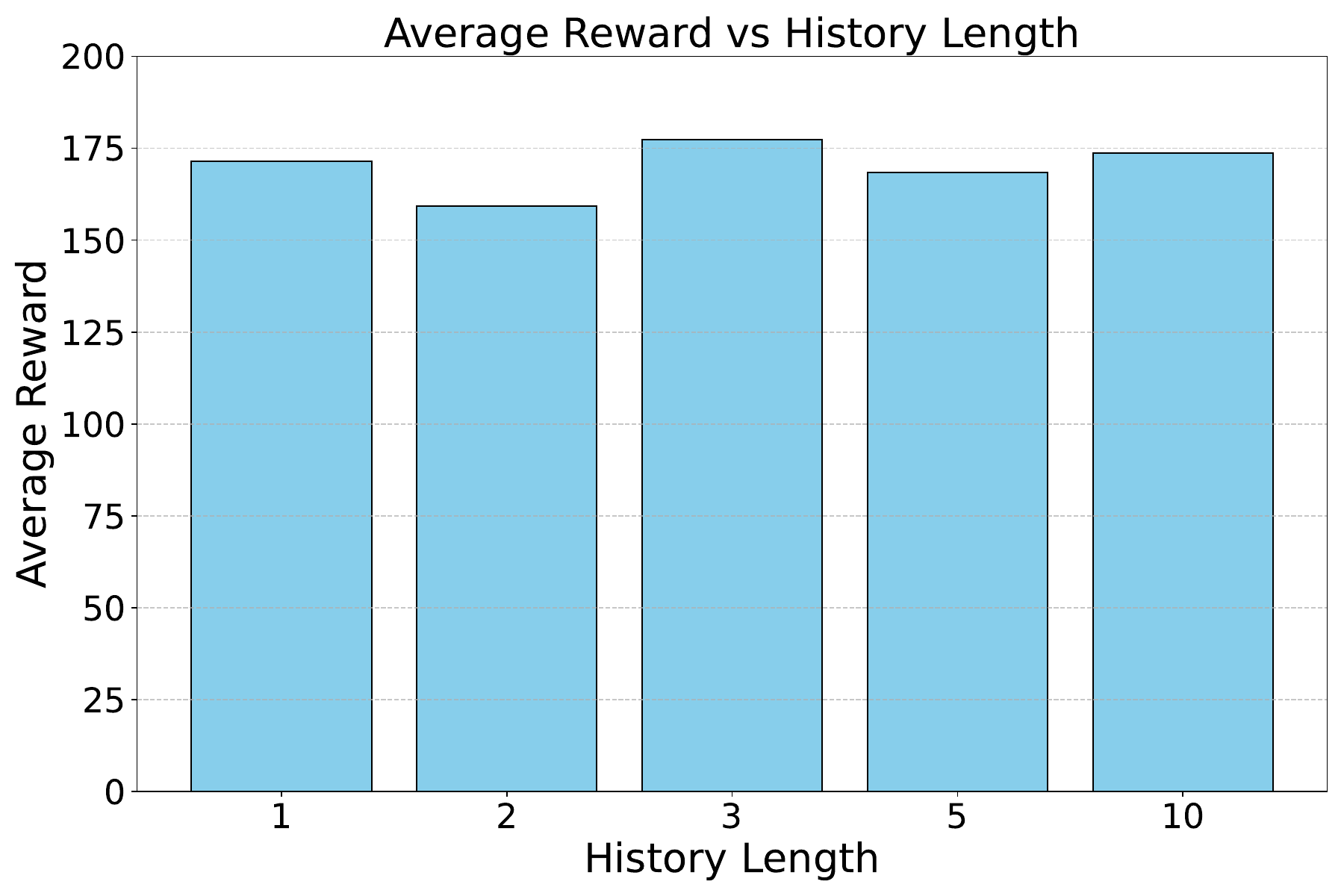}
    }{
        \caption{Mean reward of policies with different history lengths over 10 runs at test time in the Gazebo environment.}\label{fig:history_test_time}
    }
\end{floatrow}
\end{figure}

Figures \ref{fig:training_curves_history} and \ref{fig:history_test_time} show that including historical observations has little impact on performance. We hypothesize that the load's pose—even without historical data—contains enough information to estimate the other agents' states, enabling implicit communication among the MAVs. Further investigation into the role of history in more complex scenarios, such as those with higher noise or additional MAVs, is left for future work. 

\subsection{Reward function formulation} \label{appendix:reward_formulation}
The reward function components are formulated as:

\begin{equation}
\begin{aligned}
    r_t^{\mathrm{pos}} &= \lambda_1 \exp\left( -\lambda_2\norm{\bm{p}_G - \bm{p}_L} \right),\\
    r_t^{\mathrm{ori}} &= \lambda_3 \exp\left(-\lambda_4\theta(\bm{q}_{\mathrm{G}}, \bm{q}_{\mathrm{L}}) \right),\\ 
    r_t^{\mathrm{down}} &= \lambda_5 \left(1 - \exp\left(-\lambda_6 \cdot \min_i \Vert f_{\mathrm{int}}(\bm{p}_{M,i}, \bm{t}_{i}) - \bm{p}_{L} \Vert\right)\right),\\ 
    r_t^{\mathrm{act}} &= \lambda_7\exp\left( -\norm{(\bm{a}_t - \bm{a}_{t-1})/N}^2 \right),\\ 
    r_t^{\mathrm{br}} &= \lambda_8\exp\left( -\norm{\bm{\omega}_t/N} \right),\\ 
    r_t^{\mathrm{thrust}} &= \lambda_9\exp\left( -\max(\bm{T}_t/T_{\mathrm{max}}) \right),
\end{aligned}
\end{equation}

Here $\bm{p}_G$ and $\bm{p}_L$ denote the goal and load positions respectively. $\theta(\bm{q}_{\mathrm{G}}, \bm{q}_{\mathrm{L}})$ denotes the quaternion error magnitude function which is calculated using the quaternion representation of the goal orientation $\bm{q}_{\mathrm{G}}$, and the load orientation $\bm{q}_{\mathrm{L}}$. The error is calculated by taking the norm of the axis-angle representation of the quaternion difference $\bm{q}_{\mathrm{G}} \otimes \bm{q}_{\mathrm{L}}^*$, where $\bm{q}_{\mathrm{L}}^*$ is the conjugate of $\bm{q}_{\mathrm{L}}$. 

The function $\bm{f}_{\mathrm{int}}(\bm{p}_{M,i}, \bm{t}_{i})$ computes the intersection point between two elements: the line defined by the $i$-th MAV's position $\bm{p}_{M,i}$ and its thrust direction $\bm{t}_{i}$, and the plane containing the payload. This payload plane is characterized by its normal vector $\bm{n} = \bm{\ell}_x \times \bm{\ell}_y$, where $\bm{\ell}_x$ and $\bm{\ell}_y$ represent arbitrary vectors spanning the load's local $x$-$y$ plane. From all such intersection points computed for each MAV, the operator $\min$ selects the closest one to the payload position, corresponding to the most significant downwash effect.

The intersection calculation expands to:
\begin{equation}
    \bm{f}_{\mathrm{int}}(\bm{p}_{M,i}, \bm{t}_i) = \bm{p}_{M,i} + \left(\frac{d - \bm{n}\cdot\bm{p}_{M,i}}{\bm{n}\cdot\bm{t}_i}\right)\bm{t}_i
\end{equation}
where $d = \bm{n}\cdot\bm{p}_{L}$ defines the payload plane's offset from the origin through the payload position $\bm{p}_{L}$. 

The amount of MAVs is denoted by $N$, and $\bm{a}$ represents the control command, and $\bm{\omega}$ the body rate part of the control command. $\bm{T} \in \mathbb{R}^{4N}$ is the vector containing the rotor thrusts from each MAV, which is then normalized by the maximum thrust output $T_{\mathrm{max}}$. $\lambda_1, \lambda_2 \cdots \lambda_9$ are different positive hyperparameters.   All components are normalized by the simulation frequency. The chosen hyperparameters are shown in Table \ref{tab:reward_weights}.

\subsection{Training configuration} \label{appendix:training_config}

The inputs to the network are normalized stacked observation histories with history size $H=3$. We also implement a form of advantage filtering~\cite{cusumano2025robust} where $50\%$ of the samples with the lowest advantage magnitude are dropped. This approach prioritizes learning from the most informative state transitions—specifically the underexplored extremes of the data distribution where actions have a clearly better or worse outcome—thereby improving data efficiency during training. For a complete overview of the network and agent parameters, we refer the readers to Table \ref{tab:hyperparams_MAPPO}.

For setups with more than 3 MAVs, the mass of the load is sampled from a uniform distribution between 1.0 and 1.8 kg (the mass of the real payload is 1.4 kg). For the 3-MAV setup, the cables are modeled as rigid rods of 1 meter in length, connected to both the payload and the MAVs via ball joints. When using more than 3 MAVs, the system becomes overconstrained, which can lead to cable slack~\cite{fink_planning_2011}. To address this, the cables are instead modeled as three rigid segments linked by ball joints.

The episodes have a duration of 20 seconds, where a single goal pose is given to encourage stable hovering of the payload. The episode times out after 20 seconds, in which case the return is bootstrapped using the value function estimate, or it terminates earlier if:
\begin{itemize}
    \item any MAV or the payload is too close to the ground,
    \item the angle between the payload and the cable exceeds a certain threshold,
    \item the angle between the cable and the MAV exceeds a certain threshold,
    \item cables collide with each other,
    \item MAVs collide with each other,
    \item any rigid body is outside a specified bounding box,
    \item any of the cable tensions are below a specified threshold. ($>3$ MAVs)
\end{itemize}

\textbf{Reward function weights} The reward function weights shown in Table \ref{tab:reward_weights} are based on iterative tuning in simulation and real-world experiments.

\begin{table}[H]
    \centering
    \begin{tabular}{c|c}
        \hline
        \rowcolor{white}
        \textbf{Reward weight} & \textbf{Value} \\
        \hline
        \rowcolor{lightblue}
        $\lambda_1$ & 1.5 \\
        \rowcolor{white}
        $\lambda_2$ & 1.5 \\
        \rowcolor{lightblue}
        $\lambda_3$ & 1.5 \\
        \rowcolor{white}
        $\lambda_4$ & 1.5 \\
        \rowcolor{lightblue}
        $\lambda_5$ & 0.5 \\
        \rowcolor{white}
        $\lambda_6$ & 3.0 \\
        \rowcolor{lightblue}
        $\lambda_7$ & 0.5 \\
        \rowcolor{white}
        $\lambda_8$ & 0.5 \\
        \rowcolor{lightblue}
        $\lambda_9$ & 0.5 \\
        \hline
    \end{tabular}
    \caption{Reward function weights}
    \label{tab:reward_weights}
\end{table}

\textbf{Hyperparameters of MAPPO} The hyperparameters of MAPPO are shown in table \ref{tab:hyperparams_MAPPO}. The names of the parameters are based on the SKRL~\citep{serrano2023skrl} learning library.

\begin{table}[H]
    \centering
    \begin{tabular}{c|c}
        \hline
        \rowcolor{white}
        \textbf{Hyperparameter} & \textbf{Value} \\
        \hline
        \rowcolor{lightblue}
        number of envs & 4096 \\
        \rowcolor{white}
        rollouts & 128 \\
        \rowcolor{lightblue}
        learing epochs & 5 \\
        \rowcolor{white}
        mini batches & 4 \\
        \rowcolor{lightblue}
        discount factor & 0.99 \\
        \rowcolor{white}
        gae lambda & 0.95 \\
        \rowcolor{lightblue}
        learning rate actor & 5e-4 \\
        \rowcolor{white}
        learning rate critic & 1e-4 \\
        \rowcolor{lightblue}
        state preprocessor & RunningStandardScaler \\
        \rowcolor{white}
        shared state preprocessor & RunningStandardScaler \\
        \rowcolor{lightblue}
        value preprocessor & RunningStandardScaler \\
        \rowcolor{white}
        grad norm clip & 1.0 \\
        \rowcolor{lightblue}
        ratio clip & 0.1 \\
        \rowcolor{white}
        value clip & 0.1 \\
        \rowcolor{lightblue}
        entropy loss scale & 0.001 \\
        \rowcolor{white}
        value loss scale & 1.0 \\
        \rowcolor{lightblue}
        kl threshold & 0.0 \\
        \hline
    \end{tabular}
    \caption{MAPPO hyperparameters based on SKRL~\citep{serrano2023skrl} learning library}
    \label{tab:hyperparams_MAPPO}
\end{table}

\end{document}